\documentclass{article}
\usepackage{spconf,amsmath,graphicx}

\title{Does human speech follow Benford's Law?}
\name{Leo Hsu and Visar Berisha\thanks{[lhsu12, visar]@asu.edu}}
\address{Arizona State University\\ Tempe, AZ}
\begin{document}
\ninept
\maketitle
\begin{abstract}
  Researchers have observed that the frequencies of leading digits in many man-made and naturally occurring datasets follow a logarithmic curve, with digits that start with the number 1 accounting for $\sim 30\%$ of all numbers in the dataset and digits that start with the number 9 accounting for $\sim 5\%$ of all numbers in the dataset. This phenomenon, known as Benford's Law, is highly repeatable and appears in lists of numbers from electricity bills, stock prices, tax returns, house prices, death rates, lengths of rivers, and naturally occurring images. In this paper we demonstrate that human speech spectra also follow Benford's Law on average. That is, when averaged over many speakers, the frequencies of leading digits in speech magnitude spectra follow this distribution, although with some variability at the individual sample level. We use this observation to motivate a new set of features that can be efficiently extracted from speech and demonstrate that these features can be used to classify between human speech and synthetic speech. 
\end{abstract}
\vspace{0.3cm}

\section{Introduction}

   Researchers have observed that the frequencies of leading digits in many man-made and naturally occurring datasets follow a logarithmic curve, with digits that start with the number 1 accounting for $\sim30\%$ of numbers and digits that start with the number 9 accounting for $\sim5\%$ of all numbers. This phenomenon is known as {\em Benford's Law}; for lists that follow Benford's Law, the probability that the first digit in a list of numbers is $d$, $P(d)$, is

\begin{equation}
	P(d)=log_{10}(1 + \frac{1}{d}).
\end{equation}

This phenomenon has been empirically observed in datasets such as stock prices, tax returns, house prices, death rates, naturally occurring images, etc. Common across these datasets is that they span multiple orders of magnitude; when the data generating process can be modeled as the product of multiple independent factors, datasets tend to follow Benford's Law [1, 2, 3].

Benford’s law is highly repeatable across many datasets that span multiple orders of magnitude. As a result, one of its principal applications is in detection of fraud. For example, Benford’s law can be used to detect multiple compression of images in forensics applications. In images compressed using JPEG, the Discrete Cosine Transform (DCT) coefficients follow Benford’s law [4]. If the coefficients abnormally deviate from the expected distribution, the implication is that the image may have been tampered with (e.g. saved multiple times, modified artificially, etc.) Other applications of Benford's Law include detection of scientific fraud [5], e-commerce price distributions [6], macroeconomics [7], etc.

While scientists have shown empirically that many naturally-occurring data sets follow Benford’s law, to the best of our knowledge, none have analyzed whether human speech follows this phenomenon. The source-filter paradigm models speech as the output of a series of linear time-invariant filters [8]. In the frequency domain, this corresponds to the product of multiple spectra, the same conditions  that tend to produce datasets that adhere to Benford's Law  [1]. Therefore, this observation leads to the hypothesis that speech magnitude spectra will also follow Benford's Law. In this paper, we analyze the speech magnitude spectra of several hundreds speakers from a phonemically-balanced corpus to determine whether the leading digit follows the distribution in Eqn. (1); we use the results of this analysis to propose a new representation for speech.  The principal contributions of this work are as follows:
\begin{itemize}
\item We demonstrate empirically that human speech spectra follow Benford’s Law on average, with some variability at the individual sample level
\item We propose a new set of speech features based on Benford’s Law which can be easily extracted from speech and provide empirical evidence that these features can classify between human speech and synthetic speech
\end{itemize}

%The remainder of this paper is organized as follows. First, we discuss the computational model used to determine whether human speech follows Benford's Law. Next, we propose a new set of features, dubbed the Benford similarity (BenS) features, that can be extracted with little computational burden to characterize the {\em Benfordness} of speech. In the results section, we confirm that speech follows Benford's Law and empirically demonstrate the utility of BenS features for classifying between human speech and synthetic speech. We end the paper with a discussion of why speech follows Benford's Law and other potential applications of the BenS features.

\section{Methods}

In this section we provide an overview of the methodology to evaluate whether speech follows Benford's Law (Section 2.1) and propose a new set of speech features based on Benford's Law (Section 2.2).

\subsection{Evaluating whether speech follows Benford's Law}

%\begin{figure}[h!]
%    \centering
%    \includegraphics[width=3.1in]{BenfordSpeech.pdf}
%    \caption{A block diagram of the approach used to evaluate whether speech follow's Benford's Law. Input speech samples are processed frame-by-frame to compute the short-time Fourier transform. We estimate the frequency of leading digits of the  magnitude spectrum for each frame and average over all frames from all speakers. This empirical probability distribution is compared against the Benford distribution using a linear regression model.}
%    \label{fig:my_label}
%\end{figure}

The first part of Fig.1 (the red blocks) provides a block diagram of the frame-based analysis pipeline used to determine whether speech follows Benford's Law. The TIMIT train database is used for analysis [9]. The TIMIT train set contains audio recordings from 326 male and 136 female speakers; in this corpus, each speaker produced 10 sentences - 2 sentences common to all speakers and 8 non-overlapping sentences. We use this data to determine whether the speech magnitude spectrum follows Benford's Law by comparing the probabilities in Eqn. (1) with empirical estimates derived from the frequency of leading digits in the speech magnitude spectrum. Prior to analyzing the audio, dithering was used to randomize quantization errors that appear in the audio, with noise energy set at $1/1000$ of peak speech signal energy. The dithering had no perceptual impact on speech quality.
%That is, we intentionally add a small amount of noise to remove trails of zeros that were the result of quantization in the data. The noise was Gaussian distribution with mean 0 and variance determined by dividing the maximum value of the speech signal by 1000. 

Our working hypothesis was that speech magnitude spectra follow Benford's Law. To evaluate this, we compute the short-time Fourier transform of the signal with a 25ms frame length and 10ms overlap. For each frame, the DC component of the speech signal was removed by subtracting the mean from the speech signal frame prior to Fourier analysis. We remove the first element of the magnitude spectrum from consideration since it is zero (after de-meaning). We scale the magnitude spectrum of the frame by the inverse of the smallest value to ensure all values are greater than 1. The leading digits of the resulting vector are extracted to estimate their respective probabilities via a frequency histogram. This process is repeated for all frames in a speech signal; then across all audio samples from all speakers in the TIMIT database. 

We calculate the probability distribution at the speaker level by normalizing the frequency histogram by the total number of frames for that speaker. This results in an empirical estimate of the distribution of the leading digit of spectral magnitudes; this distribution is then averaged across all speakers. 

We compare the resulting average across all speakers with the ideal Benford distribution using a linear statistical test (linear regression). The linear model provides an easy-to-interpret means of evaluating our hypothesis that speech spectra follow Benford's law. A match between these two sets of probabilities will result in a linear model with a slope of 1 and a $y$-intercept of 0. 

\subsection{Benford Similarity  (BenS) Features}

As we will present in Section 3, when averaged over multiple  speakers, speech magnitude spectra follow Benford's Law. This finding motivates a new feature set for characterizing the {\em Benfordness} of a given speech sample on frame-by-frame basis. Briefly, this feature set is based on estimates of the leading digits' probabilities for a single speech frame (using the same pipeline as in the previous section), a comparison of the empirical distribution for a frame to the ideal Benford distribution, and extraction of several statistics from these comparisons across a complete sample. A high-level block diagram of the feature extraction steps are shown in Fig. 1.

\begin{figure}
    \centering
    \includegraphics[width=3.1in]{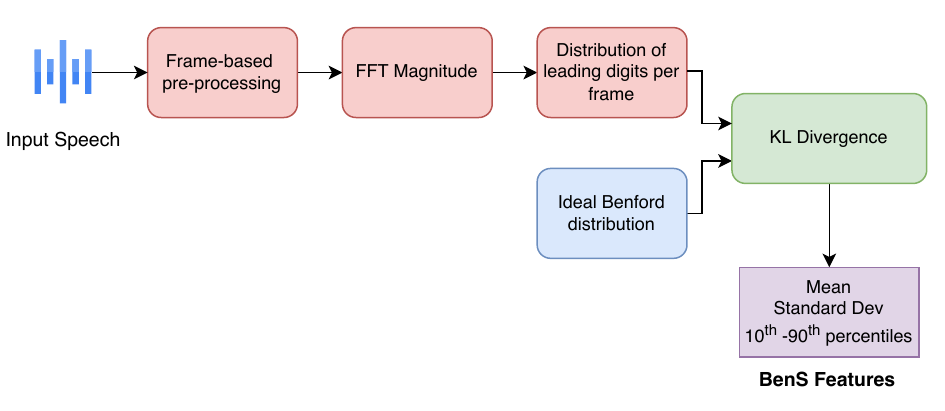}
    \caption{A block diagram of the approach used to extract the BenS Features. Input speech samples are processed frame-by-frame to compute the short-time Fourier transform. The leading digits of the spectral magnitude for each frame is used to estimate the frequency of leading digits in the magnitude spectrum. We derive a continuous feature from the distribution by comparing it against the ideal Benford distribution using the Kullback-Leibler (KL) divergence. This process is repeated for every frame of an input sample; 11 statistics are then computed from the resulting KL divergence values.}
    \label{fig:my_label2}
\end{figure}

To extract the speech features, the first step follows the frame-based pipeline described in the previous section. The leading digits are used to estimate the empirical Benford distribution, $\hat{P}(d)$, which is compared against the ideal Benford Distribution, $P(d)$, using the Kullback-Leibler (KL) Divergence,

\begin{equation}
    D_{KL}(P||\hat{P}) = \sum_{d=1}^{9} P(d)log_{10} \left ( \frac{P(d)}{\hat{P}(d)} \right ).
\end{equation}

The KL divergence evaluates the similarity between the two probability mass functions; a small KL Divergence implies that the empirical distribution is very similar to the ideal Benford distribution whereas a large KL divergence implies that the distributions are very different. 

For each frame, we obtain a single feature, the KL divergence between the ideal and the empirically-estimated distributions. This process is then repeated across all frames in a sample. For some frames, the empirical estimate of the less frequent numbers in the Benford distribution are 0, which makes it impossible to estimate the KL Divergence (since it goes to $\infty$). As a result, we remove all frames which have $0$ estimates for any of the empirical probabilities. The distribution of the remaining KL divergence values are characterized using 11 statistics; these form the final feature set extracted at the sample level. These statistics include: (1) mean of the KL, (2) standard deviation of the KL, (3)-(11) the $10^{th}$-$90^{th}$ percentiles of the KL. We refer to this feature set as the Benford Similarity  (BenS) Features.

\section{Results}

In this section we present the results of our analysis. In section 3.1, we provide evidence that speech follows Benford's Law. In section 3.2, we empirically evaluate the utility of the BenS features to distinguish between human speech and synthetic speech.

\subsection{Does human speech follow Benford's Law?}

Evaluating individual magnitude spectra sample-by-sample (or speaker-by-speaker) for Benfordness, we notice non-trivial variability in the empirical distributions of the leading digits (more on this in the next section). However, when averaged over all speakers in TIMIT train, the results are clearer. In Fig. 2, we compare the probabilities of the ideal Benford distribution with the empirical probabilities estimated by averaging across all speakers in the TIMIT train corpus, per section 2.1. It's obvious from visual inspection of this figure that there is strong agreement between the empirical and ideal distributions. We use a linear regression between empirical ($E$) and ideal ($I$) Benford probabilities to evaluate our original hypothesis; the analysis yields the equation
\begin{equation}
    E = 1.046 I - 0.005,
\end{equation}
which was significant with $p<0.05$. A slope near 1 and an intercept near 0 provides further evidence that speech spectra follow Benford's Law when averaged across the hundreds of speakers available in TIMIT. The association is confirmed statistically as the model has $R^2=0.99$.

\begin{figure}
	\includegraphics[width=\linewidth]{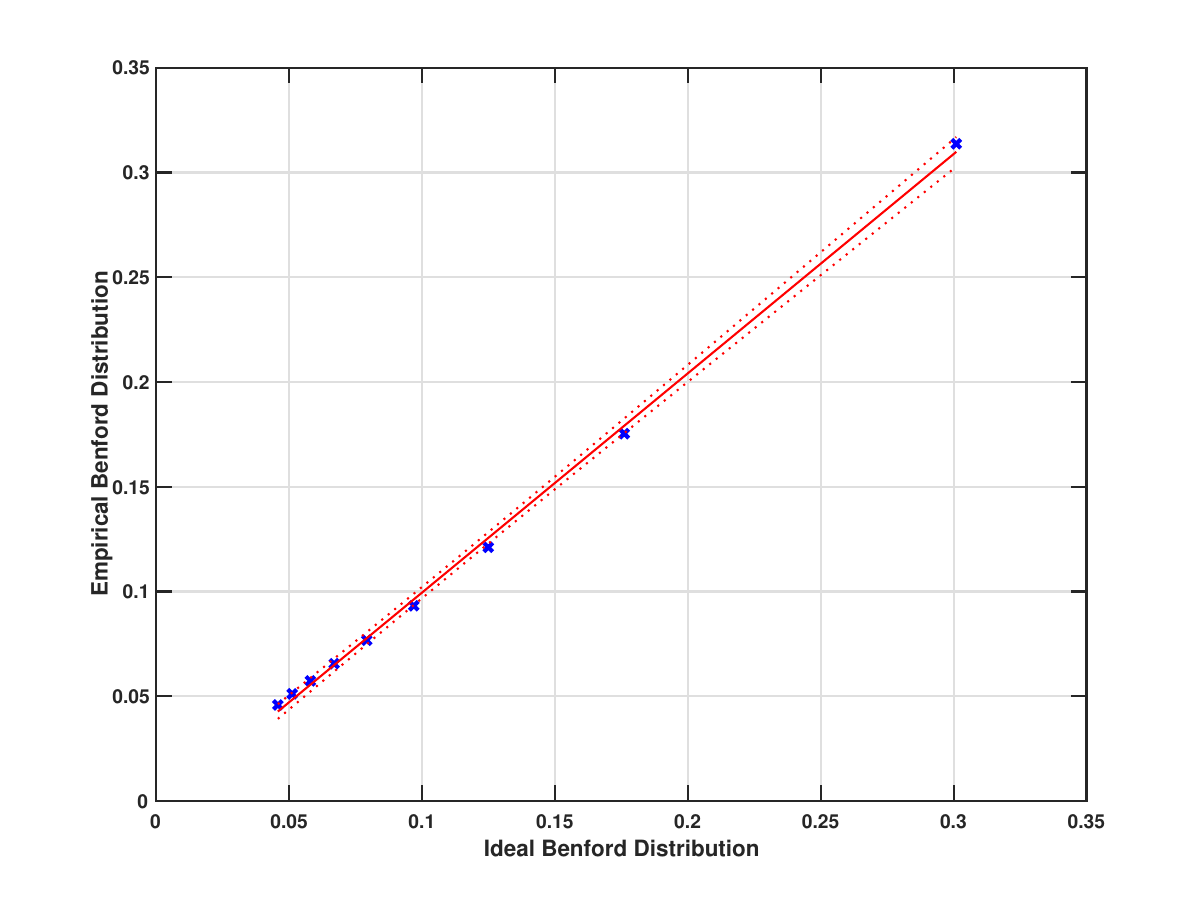}
	\centering
	\caption{This figure shows the agreement between the empirical and ideal Benford distribution probabilities for human speech (averaged over the TIMIT. A linear model fit between the empirical ($E$) and ideal ($I$) Benford distributions yields the equation  $E = 1.046 I - 0.005$. A slope near 1 and an intercept near 0 provides strong evidence that the empirical distribution matches the ideal Benford distribution. The model has $R^2=0.99$.}
	\label{fig:fig1}
\end{figure}

\subsection{Using BenS features to detect synthetic speech}

{\bf Dataset generation:} With deepfake technology becoming increasingly more sophisticated, new techniques are required to distinguish human speech from synthetic speech. To that end, we evaluate whether BenS features can be used to classify between real speech samples and synthetic samples.

We consider 20 different US-English voices from male speakers and female speakers from Google’s text-to-speech API. The Google API includes 10 TTS voices based on a pre-trained Wavenet model with a high mean opinion score (MOS) and 10 standard voices based on other models [10]. We compare speech from these synthetic voices to 20 speakers from TIMIT. We match the gender and the  content of the spoken text across the two groups by selecting 20 TIMIT speakers from the TIMIT DR2 test set. We used the 10 sentences spoken by each of the 20 human TIMIT speakers to generate matched samples from the 20 TTS voices. We extracted the BenS features from each of the 10 sentences spoken by 20 human speakers and 20 synthetic speakers. This results in two data sets, each consisting of 200 samples (20 speakers $\times$ 10 samples/speaker); from each sample we extract the 11 BenS features. We match the content and gender of each sample across the two classes in an effort to reduce potential confounding effects in classification.

\vspace{0.2cm}
\noindent {\bf Classification experiment:} The two datasets were used to train a classifier to distinguish between human speech and synthetic speech. We evaluate the model using leave-one-speaker-out (LOSO) cross-validation. That is, we remove one speaker from the training set and train the model using the data from the remaining speakers. After training the model, the data from the removed speaker is used to evaluate the accuracy of the model. This is repeated across all speakers, one by one, and the resulting estimate of accuracy is considered the out-of-sample estimate of the model’s accuracy. We considered a linear support vector machine (SVM), a decision tree, and several higher-order polynomial-kernel SVMs. While all models performed reasonably well, the best performing model was the quadratic SVM (QSVM) with an accuracy of 91.5 percent and an error rate of 8.5 percent. Table 1 shows the confusion matrix of the QSVM model. 

It’s clear from the results that the BenS features separate human speech from synthetic speech, albeit with some error. That is, there is a difference in the distribution of BenS features between human speech and speech generated by existing TTS systems. We can visualize this difference in distributions between human speech and synthetic speech via a scatter plot of the first two BenS features, the mean and standard deviation of the KL divergence values estimated for a given sample. In Fig. 3, we plot these two features for the 200 samples from the 20 human speakers (blue) and the 200 samples from the 20 synthetic speakers (red). The figure shows the separability in the two features between the groups.

\begin{figure}
	\includegraphics[width=\linewidth]{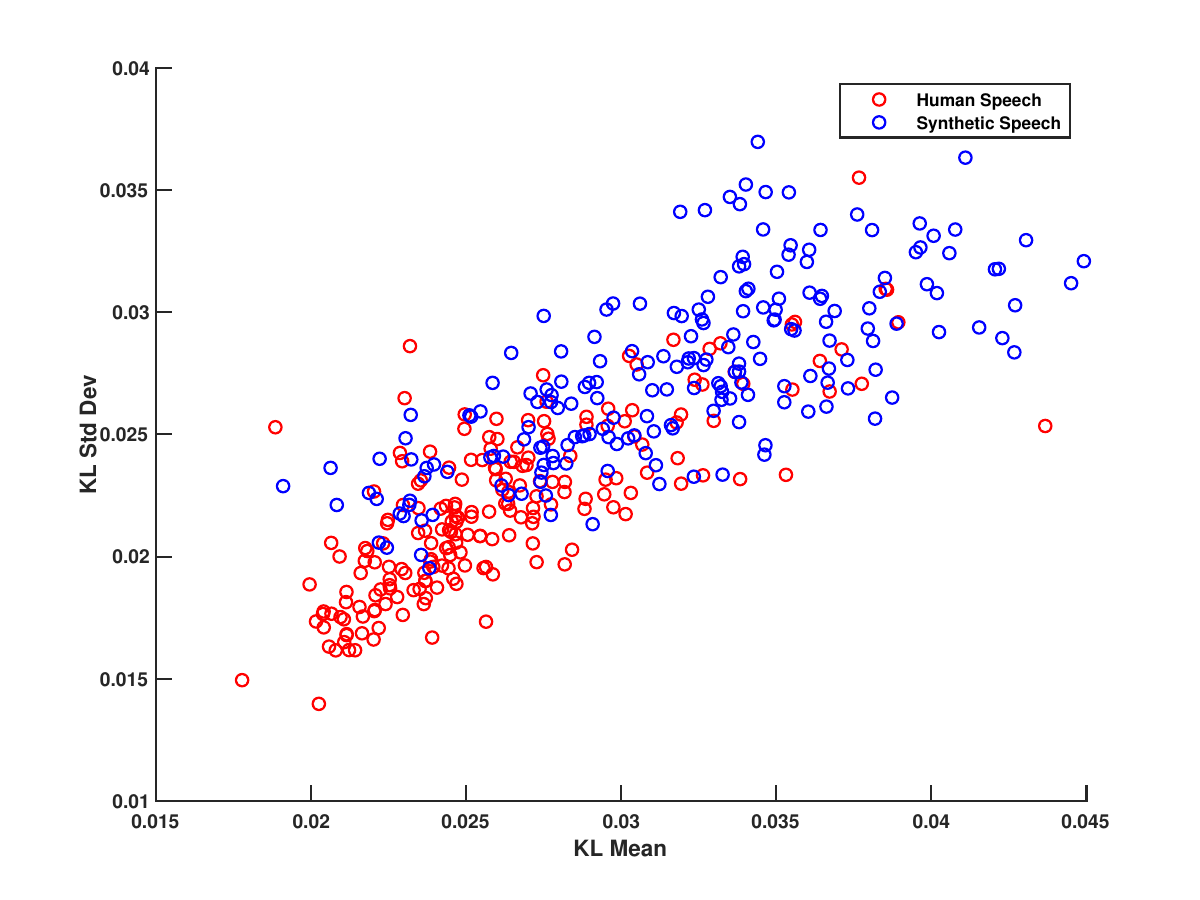}
	\centering
	\caption{A scatter plot of two BenS features, the mean of the KL divergence and the standard deviation of the KL divergence, for human speech and for synthetic speech. There is increased variability across both feature dimensions for synthetic speech.}
	\label{fig:fig2}
\end{figure}

\begin{table}
\centering
\caption{The confusion matrix for the classifier that uses the BenS features to distinguish between synthetic speech and human speech. The matrix was computed using leave-one-speaker-out cross-validation.  }
\begin{tabular}{c|ll}

             & \multicolumn{2}{c}{Predicted Classes} \\ \hline
True Classes & \multicolumn{1}{c}{Human} & Synthetic \\ \hline
Human        & \multicolumn{1}{c}{181}   & 19        \\ 
Synthetic    & \multicolumn{1}{c}{15}    & 185       \\ 
\end{tabular}
\end{table}

\vspace{0.2cm}
\noindent {\bf Individual BenS feature analysis:} While the classification results in the previous section are promising, the sample size is comparatively small due to the small number of available deepfake voices. For feature interpretability, we use the TIMIT training data used to confirm that speech follows Benford’s law to normalize the features extracted from the 20 test human speakers and the 20 synthetic speakers. That is, the mean and standard deviation for each of the 11 BenS features from the TIMIT training set was used to $z$-score the features from the data used in the classification experiment, 
\begin{equation}
	z_i=\frac{x_i-\mu_i}{\sigma_i},
\end{equation}
where $x_i$ is the $i^{\mathrm{th}}$ BenS feature, $\mu_i$ and $\sigma_i$ are the mean and standard deviation of that feature (estimated from the TIMIT training set), and $z_i$ is the normalized version of that feature. 

Table 2 shows the $z$-scores for each feature extracted from human and synthetic speakers. It is clear that there are differences between the features for human and synthetic speech. The human speech samples extracted from the 20 human speakers have $z$-scores within approximately 1 standard deviation of the feature distributions estimated from the larger TIMIT training set. In contrast, the  synthetic speech has $z$-scores that exceed 1 standard deviation from the mean. On the one hand, this finding makes sense as we would expect that synthetic speech samples are distinct from human speech given the results of the classification analysis. On the other hand, the $z$-scores for the 20 human speakers are not zero; rather, they are positively biased to approximately $\sim$0.8 standard deviations from the mean. We posit that this bias is a result of a mismatch in the data distributions between the complete TIMIT training set and the small subset of speakers we used from the TIMIT test set. Our TIMIT test set contained speakers from only one dialect region (DR2), whereas the TIMIT training set contains data from all dialect regions. We controlled for the dialect region in our test set in an effort to reduce variability (and control for potential confounders) for our classification analysis given the small sample size. The sample size was limited given the small number of synthetic voices available.

\begin{table}
\centering
\caption{$z$-scores for the BenS features for human and synthetic speech.}
\begin{tabular}{cll}
\hline
\multicolumn{1}{c}{Feature}            & \multicolumn{1}{c}{Human Speech}    &\multicolumn{1}{c}{Synthetic Speech}  \\ \hline
\multicolumn{1}{c}{mean}               & \multicolumn{1}{c}{0.859}                       & \multicolumn{1}{c}{1.861}   \\                                                     
\multicolumn{1}{c}{standard deviation} & \multicolumn{1}{c}{0.814}                       & \multicolumn{1}{c}{1.812}   \\ 
\multicolumn{1}{c}{10th Percentile}    & \multicolumn{1}{c}{0.784}                       & \multicolumn{1}{c}{1.409}   \\
20th Percentile                          & \multicolumn{1}{c}{0.814}                       & \multicolumn{1}{c}{1.379}   \\
30th Percentile                          & \multicolumn{1}{c}{0.814}                       & \multicolumn{1}{c}{1.401}   \\ 
40th Percentile                          & \multicolumn{1}{c}{0.847}                       & \multicolumn{1}{c}{1.423}   \\ 
50th Percentile                          & \multicolumn{1}{c}{0.850}                       & \multicolumn{1}{c}{1.579}   \\ 
60th Percentile                          & \multicolumn{1}{c}{0.856}                       & \multicolumn{1}{c}{1.915}   \\ 
70th Percentile                          & \multicolumn{1}{c}{0.874}                       & \multicolumn{1}{c}{2.320}   \\ 
80th Percentile                          & \multicolumn{1}{c}{0.892}                       & \multicolumn{1}{c}{2.358}   \\ 
90th Percentile                          & \multicolumn{1}{c}{0.874}                       & \multicolumn{1}{c}{1.966}   \\ \hline 
\end{tabular}
\end{table}

\vspace{0.2cm}
\noindent {\bf Variability in the {\em Benfordness} of speech:} The results of Section 3.1 provide evidence that human speech follows Benford's law when averaged over hundreds of speakers. However, Fig. 3 demonstrates that there is sample-to-sample variability in the Benfordness of speech. If the speech of a speaker follows Benford's law, then the KL divergence between the empirically-estimated distribution and the ideal Benford distribution should be 0. However, as we see from the distribution of the human speech in Fig. 3 (red samples), there is some variability in the mean of the KL divergence from speaker to speaker. In fact, as the scatter plot in Fig. 3 shows, variability plays an important role in  distinguishing between human and synthetic speech.

\section{Discussion}

The evidence in Fig. 2 and the results of the  statistical analysis suggest that speech spectra follow Benford's Law on average. 
It is important to note that there is a non-trivial amount of variability in the frequency of leading digits in individual speech magnitude spectra. That is, while on average the leading digits in speech magnitude spectra followed Benford's Law, this was not true for all speech spectra individually. To more completely characterize the Benfordness of speech, we developed a new  multi-dimensional representation of Benford similarity that is comprised of the mean and several measures of variability.

The fact that speech observes Benford's law on average is not surprising when we consider existing models of the human speech production mechanism. Benford’s Law is apparent in datasets where the data generating process is the product of multiple independent factors [1]. Source-filter theory models speech as the output of a sequence linear time-invariant systems at the frame level [8]. That is, in the frequency domain, we  model the speech spectrum $S(\omega)$ as,

\begin{equation} \label{eqn:speechproduction}
S(\omega) = E(\omega) V(\omega) R(\omega),
\end{equation}
where $E(\omega$ is the glottal excitation signal in the frequency domain, $V(\omega)$ is the spectrum of the impulse response of the vocal tract filter, and $R(\omega)$ is the spectrum of the impulse response of a filter that models the radiation characteristics at the mouth. The equation above models speech as the product of multiple factors. While $E(\omega)$, $V(\omega)$, and $R(\omega)$ are not completely independent, most instantiations of the source-filter model make the simplifying assumption that they are. Adherence to Benford’s law provides additional evidence that this independence assumption is justified as Benford’s Law holds for datasets generated via a product of multiple {\em independent} factors [1]. 

We further demonstrated that the BenS features are useful to classify between human and synthetic speech. The distribution of KL divergence values over a sample – as measured by the BenS features – were markedly different for human speech and synthetic speech. The differences in the BenS features are perhaps not surprising when we consider differences in the data generating process between human speech and synthetic speech. Human speech is modeled as the output of Eqn. (5); however the most sophisticated synthetic voices used in our database were produced by neural networks with complex non-linear structures. It’s possible that the BenS features capture these fundamental differences in the underlying mechanism of production; however, evaluating this hypothesis requires a deeper study of synthetically-generated speech.

While in this paper we provide evidence that the features are useful for detection of synthetic speech, other applications of BenS features may be possible. For example, for speech produced by patients with certain clinical conditions, the independence assumption between $E(\omega)$ and $S(\omega)$ is more difficult to justify. In neurological conditions where velopharyngeal control is impacted (e.g. amyotrophic lateral sclerosis [11]), air escapes through the nasal cavity during speech (changing $V(\omega))$, therefore the speaker may actively attempt to increase the loudness of their speech (thereby changing $E(\omega)$). This clearly makes $E(\omega)$ and $V(\omega)$ dependent on each other. We conjecture that this would lead to changes in the BenS representation. As a result, follow-on work should evaluate the utility of this representation for detecting early changes in clinical conditions via speech. If validated, the computational simplicity of the feature set would make it appealing for persistent on-device tracking. 

\section{Conclusion}

We evaluated the distribution of leading digits in the speech spectrum (extracted at the frame level) and showed that it follows Benford’s law on average. One possible explanation for adherence to Benford’s law comes from existing models of the speech production mechanism. Human speech can be modeled as the product of multiple independent signals in the frequency domain and it’s known that datasets that can be modeled in this way will tend to follow Benford's Law. Based on this observation we proposed a new feature set, dubbed the Benford similarity (BenS) features, for characterizing the {\em Benfordness} of speech at the level of an utterance using the KL divergence between the empirically-estimated Benford distribution and the ideal Benford distribution. Classification results showed that this feature set can distinguish between human speech and  synthetic speech with accuracy exceeding 90\%. It’s important to note that the classification experiment was limited as the dataset consisted of only 20 speakers from the TIMIT test set. This was done in order to match the human data to the unique synthetic voices, of which we only had 20. Future work will focus on analyzing the utility of BenS Features in other contexts.

\bibliography{mybib}

\end{document}